# Leibniz's Monadology as Foundation for the Artificial Age Score: A Formal Architecture for AI Memory Evaluation

Seyma Yaman Kayadibi

Victoria University

seyma.yamankayadibi@live.vu.edu.au

## 1. Abstract

This paper develops a mathematically rigorous, philosophically grounded framework for evaluating artificial memory systems, rooted in the metaphysical structure of Leibniz's Monadology. Building on a previously formalized metric, the Artificial Age Score (AAS), the study maps twenty core propositions from the Monadology to an information-theoretic architecture. In this design, each monad functions as a modular unit defined by a truth score, a redundancy parameter, and a weighted contribution to a global memory penalty function. Smooth logarithmic transformations operationalize these quantities and yield interpretable, bounded metrics for memory aging, representational stability, and salience. Classical metaphysical notions of perception, apperception, and appetition are reformulated as entropy, gradient dynamics, and internal representation fidelity. Logical principles, including the laws of non-contradiction and sufficient reason, are encoded as regularization constraints guiding memory evolution. A central contribution is a set of first principles proofs establishing refinement invariance, structural decomposability, and monotonicity under scale transformation, aligned with the metaphysical structure of monads. The framework's formal organization is structured into six thematic bundles derived from Monadology, aligning each mathematical proof with its corresponding philosophical domain. Beyond evaluation, the framework offers a principled blueprint for building AI memory architectures that are modular, interpretable, and provably sound.

## 2. Introduction

The Artificial Age Score (AAS) framework (Kayadibi, 2025) is extended by a theoretically grounded scaffold based exclusively on Leibniz's Monadology. Rather than seeking patterns purely from data, an existing mathematical core is taken as given and strengthened with constraints distilled from classical metaphysics, such that Monadology is treated as a design blueprint for information-theoretic memory systems in which boundedness, modularity, and law-governed behavior are engineered and justified (Leibniz, 1714/1948). In Leibniz's metaphysics, a monad is described as a simple, partless substance (§§1–3) that is not directly affected by external causes (§7) and yet represents the universe from its own point of view (§§56–60). Change is taken to proceed from an internal principle of appetition (§§14–15), degrees of clarity in perception are allowed (§§21–23), and "windowless" perspectives are coordinated by pre-established harmony (§7, §§78–81). In this account, these features are operationalized: module autonomy is motivated by inwardness; coordination without leakage is guided by harmony; and invariance under artificial refinements is supported by simplicity. The AAS framework (Kayadibi, 2025) is retained as the core analytical object; its primitives and diagnostics are preserved, and their Leibnizian grounding is made explicit.

### 2.1 Background and Theoretical Context

The project is grounded in the Artificial Age Score (AAS) framework (Kayadibi, 2025) and in Leibniz's Monadology (1714/1948). Clause-level constraints distilled from Monadology are compiled into AAS as enforceable modules while the AAS primitives are left intact. The penalty kernel is Shannon-consistent: degradation in recall quality is measured by a negative logarithmic surprisal cost that is zero at perfect recall, and the mapping from recall scores to probabilities, as well as any smoothing parameters, are treated as modeling choices (Shannon, 1948). Within this division of roles, Leibniz supplies the architectural constraints, AAS supplies the measurable core, and Shannon supplies the information-theoretic justification. This continuity-preserving role is aligned with Leibniz's law of continuity and his account of appetition (Leibniz, 1714/1948, §§10–12, 14–15, §56).

In the literature on artificial memory evaluation, two tracks are often distinguished: an information-theoretic track, in which penalties derived from surprisal or coding costs are used to obtain interpretable and bounded behavior (Shannon, 1948), and an architectural track, in which modularity, invariance, and auditability are emphasized to enable governance at scale while preserving the classic episodic semantic distinction (Tulving, 1972, 1985). A different starting point is adopted here: Monadology is treated as a design blueprint whose motifs are translated into enforceable engineering constraints that are typically left implicit in contemporary scoring systems, while the measurable core follows the AAS formulation (Kayadibi, 2025). Concretely, simplicity is invoked to motivate refinement invariance: artificial partitions, tokenizations, and storage layouts should not alter scores when underlying content is preserved (Leibniz, 1714/1948, §§1–3). Windowlessness together with pre-established harmony motivates causal insulation: each channel's contribution is constrained to depend only on its own history, and any cross-channel effects are routed solely through an explicit overlap term (Leibniz, 1714/1948, §7, §§78–81). Appetition is used to justify penalty-reducing, rate-limited updates, such that smooth trajectories are favored over volatile spikes (Leibniz, 1714/1948, §12, §§14–15, §56). While the AAS framework is grounded in Leibnizian monads and placed in dialogue with Kantian apperception as background, the resemblance is structural rather than doctrinal. In Leibniz, unity is grounded in simple substances that represent the world from within (Leibniz, 1714/1948, §§1–3, 7, 14–15, 56–60, 78–81), whereas in Kant, it is located in the subject's synthetic act of apperception as a transcendental condition of possible experience (Friedman, 1992; Kant, 1998, B132–B136). In the AAS model, modular memory contributions are encoded so that local autonomy is preserved and global coherence is maintained.

## 2.2 Contributions

**(1) Clause-level formalization:** Twenty propositions from the Monadology are compiled into a compact, testable clause set that (i) preserves refinement and embedding invariance; (ii) enforces causal insulation with explicit redundancy accounting; (iii) separates perception (liveness) from apperception (distinctiveness) via contribution shares and a single, auditable gate; and (iv) adds governance rules for contradiction, sufficient reason, cross-view harmony, hierarchical organicity, and AAS-specific, windowed net-drift promotion/rollback (Kayadibi, 2025; Leibniz, 1714/1948; Shannon, 1948). **(2) Information-theoretic kernel with Leibnizian grounding:** The AAS uses an ε-regularized surprisal kernel introduced in Kayadibi (2025) and defined in 3.1, which yields a bounded and interpretable penalty. The kernel's strict monotonicity and convexity follow from its logarithmic form, while the information-theoretic rationale derives from Shannon (1948). The ε-regularization and finite cap are AAS modeling choices (Kayadibi, 2025). Smoothing and rate-limited updates are conceptually grounded in Leibniz's law of continuity and appetition (Leibniz, 1714/1948). **(3) Scalable governance for memory systems:** It is demonstrated that monad-informed constraints yield diagnostics and policies that resist schema gaming, maintain a single, auditable gate between liveness and apperception, and enable cross-view release guarantees without pipeline coupling (Leibniz, 1714/1948). The underlying penalty kernel and redundancy-adjusted weighting are adopted from the AAS formulation (Kayadibi, 2025) with an information-theoretic justification (Shannon, 1948).

## 2.3 Purpose of the Study

This study is founded on the Artificial Age Score (AAS) formula (Kayadibi, 2025) and on a clause-level reconstruction of twenty propositions from the Monadology (Leibniz, 1714/1948). Its purpose is to integrate these propositions into a computable, first-principles architecture for AI memory. Each proposition is compiled as a measurable AAS clause, so that appetition, harmony, and sufficient reason are encoded, respectively, as penalty dynamics, redundancy correction, and representational constraints, and are then proved by construction. This by-construction mapping provides a bridge between classical metaphysics and contemporary AI evaluation, yielding testable and bounded laws for memory aging and stability. It also clarifies the limits of approximating ideal continuity through empirical procedures, in line with Leibniz's law of continuity (Leibniz, 1714/1898, §56, §§10–12).

### 2.4 Research Questions

**RQ1.** How can the philosophical structure of Leibnizian monads be transformed into a modular, measurable architecture for evaluating artificial memory systems with AAS?

**RQ2.** To what extent can diagnostics of memory aging and stability in large-scale AI systems be supported by a monad-informed AAS, while balancing representational clarity, redundancy, and unity under explicit information-theoretic constraints?

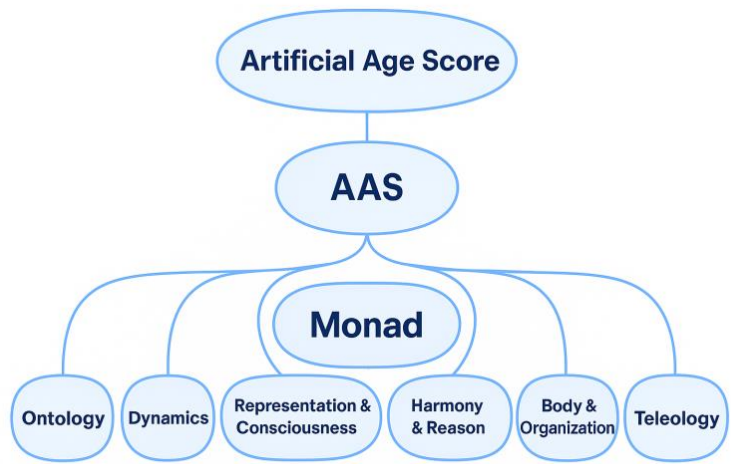

Figure 1.Artificial Age Score (AAS) to monad-like channels and six thematic bundles

### 3. Methodology

A unified evaluation pipeline is specified to formalize the metaphysical structure of the Monadology within the Artificial Age Score (AAS) framework. All ninety monads from Leibniz's Monadology (1948) are systematically read, analytically coded, and assessed for computational enforceability. From this corpus, twenty monads are retained and integrated into the AAS as enforceable modules. For clarity, they are grouped into six thematic bundles: Ontology (§§1, 3, 7, 9), Dynamics (§§10, 11, 12, 15), Representation and Consciousness (§§14, 21, 26, 29), Harmony and Reason (§§31, 32, 78, 79), Body and Organization (§§64, 70), and Teleology (§§58, 90). Clause construction follows a two-step process. First, a dominant philosophical motif is assigned to each monad to guide formalization. These include refinement, causal insulation, appetition versus peak representation, entropy, the Principle of Sufficient Reason (PSR), harmony, teleological order, and representational dominance. Second, each motif-tagged monad is formalized as a compact clause enforceable within the AAS, enabling measurable integration of metaphysical content. Each monad is structured through a philosophical premise, an operational clause, and one or more formal statements. These are mathematically derived and validated using components of the AAS model. Symbolic notation is employed, proof sketches are provided, and clause-specific diagnostics are defined where relevant, ensuring each monad functions as a testable module within a unified framework. Measurement is conducted using definitions in 3.1, including the penalty kernel, per-channel contribution, and total AAS score. For each session and channel, state variables, redundancy values, and weights are instantiated. The score is computed, and diagnostic outputs, such as contribution magnitude, distributional spread, dominance, and clause-specific penalties, are recorded. Stepwise variation is monitored to assess stability, and diagnostics are applied only when the corresponding clause is active. Validation is established analytically. Structural guarantees follow from the mathematical properties of the AAS kernel. As it is convex and decreasing, higher redundancy increases penalty while accuracy reduces it, consistent with Shannon (1948) and Kayadibi (2025). Refinement and embedding invariance are ensured by construction; any division or merging that preserves state, redundancy, and weight leaves the score unchanged. Causal insulation is maintained by restricting each channel to its own history, with any interaction routed through an explicit overlap term. Robustness holds across all admissible weights, redundancy estimators, and smoothing constants ($\varepsilon > 0$), with no tuning. Jensen-type lower bounds (§58) are applied where needed to verify order, variety, or perfection constraints. All results are derived entirely within the AAS. Each monad is supported by a self-contained proof, and thematic bundle summaries are retained to support modular interpretation and future AI architectural reuse.

## 3.1 Common Definitions

**Indexing convention.** Sessions are indexed by $t \in \{1, \ldots, T\}$ (earlier drafts used j); channels are indexed by $i \in \{1, \ldots, m\}$. Weights satisfy $w_i \geq 0$ with $\sum_i w_i = 1$. The smoothing constant satisfies $\varepsilon > 0$.

**Core score (hybrid AAS).**

**Provenance.** The hybrid Artificial Age Score (AAS) is adopted exactly as in Kayadibi (2025). Unless stated otherwise, "AAS" denotes this hybrid form.

$$AAS_t^{hyb} = \sum_{i=1}^{m} w_i \left(1 - R_{t,i}\right) \phi_\varepsilon(x_{t,i}), \text{ where } \phi_\varepsilon(x) = \log_2 \frac{1+\varepsilon}{x+\varepsilon}.$$

$$AAS_t^{hyb} = \sum_{i=1}^{m} \alpha_{t,i} \phi_\varepsilon(x_{t,i}), \text{ where } \alpha_{t,i} := w_i(1 - R_{t,i}) \geq 0, \quad (1)$$

The kernel satisfies $\phi_\varepsilon(1) = 0$ and yields a bounded, interpretable penalty: $0 \leq AAS_t \leq \phi_\varepsilon(0) = \log_2 \frac{1+\varepsilon}{\varepsilon}$.

**Per-channel contribution and session total:**

Define $S_t := AAS_t$; hence $p_{t,i} = c_{t,i}/S_t$.

$$c_{t,i} = w_i \left(1 - R_{t,i}\right) \phi_\varepsilon(x_{t,i}) \geq 0, \text{ and } S_t = \sum_i c_{t,i}. \quad (2)$$

**Apperception shares and gate.**

$p_{t,i} = c_{t,i}/S_t$ if $S_t > 0$ (else $p_{t,i} = 0$); peak share $\rho_t = \max_i p_{t,i}$; contribution entropy

$$H_t^{(contrib)} = -\sum_{i: p_{t,i} > 0} p_{t,i} \log_2 p_{t,i}.$$ A unified salience gate is $AppLevel_t = (1 - \kappa_t)\rho_t \in [0,1], \quad \kappa_t \in [0,1]. \quad (3)$

**Trajectory summaries.**

$A_i = \sum_t \alpha_{t,i}$; $time-average \; \bar{x}i = \left(\sum_t \alpha_{t,i} x_{t,i}\right)/A_i$ (if $A_i > 0$); time-spread entropy

$$H_i^{(time)} = -\sum_{t:\alpha_{t,i}>0}(\alpha_{t,i}/A_i)\log_2(\alpha_{t,i}/A_i). \quad (4)$$

**3.2 Main Analysis: Explaining Twenty Monads via AAS**
**3.2.1 Ontology (4)**
**Monad 1 - Simple Substance "without parts" (§1)**
**Philosophical premise.** In §1, the monad is described as a simple substance "without parts" that nonetheless enters into compounds (Leibniz, 1714/1948, §1).
**Operational clause.** Simplicity is operationalized as (i) refinement invariance and (ii) overlap-adjusted additivity for participation in compounds.
**Proposition 1 Refinement invariance.**
If a channel i is refined into a finite family $\{i, \alpha\}$ that preserves content and redundancy,

$x_{t,i,\alpha} = x_{t,i}$, $R_{t,i,\alpha} = R_{t,i}$ and redistributes the weight so that $\sum_\alpha w_{i,\alpha} = w_i$, then the total score given by Eq. (1) is unchanged: $AAS_t^{(refined)} = AAS_t$.

**Proof (sketch).** The score in Eq. (1) is linear in weights and depends on (x,R); since (x,R) are preserved and weights sum, the total is invariant.

**Proposition 2 Participation/mixture additivity and bounds.**

For a compound C formed from monads $M^{(j)}$ with mixture weights $\pi_j \geq 0$, $\sum_j \pi_j = 1$, let the compound-level overlap satisfy $\Lambda_{t,(j,i)} \in [0,1]$ with $\Lambda_{t,(j,i)} \geq R_{t,i}^{(j)}$. Then $0 \leq AAS_t^{(C)} \leq \sum_j \pi_j \, AAS_t^{(M^{(j)})}$, with equality when no extra inter-monad redundancy is present $\Lambda_{t,(j,i)} = R_{t,i}^{(j)}$.

**Proof (sketch).** Non-negativity follows from the kernel in Eq. (1); the upper bound follows because each term is reduced by $1 - \Lambda$, which is at most the stand-alone reduction $1-R$.

**Corollary (Diagnostic indivisibility).**

Any two channelizations of the same monad that preserve (x,R) and total weights yield identical $AAS_t$ for all t; internal "parts" are not detected by observation, matching §1.

**Monad 3 Absence of Extension, Figure, and Divisibility (§3)**

**Philosophical premise.** Where there are no parts, there is neither extension nor figure nor divisibility. In AAS, this appears as (i) independence from spatial/formal embedding and (ii) invariance under measurement-preserving refinement (Leibniz, 1714/1948, §3).

**Notation.** Core score and per-channel contribution follow §3.1, Eq. (1)–(2). Per-channel spatial/formal metadata $s_{t,i}$, used only for embeddings, is introduced.

**Definition 3.1 Embedding invariance.**

Let an embedding transform F act on metadata as $s'_{t,i} = F(s_{t,i})$. If $AAS_t$ depends only on $(w_i, R_{t,i}, x_{t,i})$ (cf. Eq. (1)), then

$$AAS_t(w_i, R_{t,i}, x_{t,i}, s_{t,i}) = AAS_t(w_i, R_{t,i}, x_{t,i}, s'_{t,i}).$$

**Lemma 3.1 No extension, no figure.**

Since $s_{t,i}$ does not appear in Eq. (1), any embedding F leaves $AAS_t$ unchanged. □

**Definition 3.2 Measurement-preserving refinement.**

Refine channel i into $\{(i, \alpha)\}$ such that $\sum_\alpha w_{i,\alpha} = w_i$, $\quad x_{t,i,\alpha} = x_{t,i}$, $\quad R_{t,i,\alpha} = R_{t,i}$.

**Lemma 3.2 No divisibility.**

Under any such refinement,

$$\sum_{i,\alpha} w_{i,\alpha}(1 - R_{t,i,\alpha})\phi_\varepsilon(x_{t,i,\alpha}) = \sum_i w_i(1 - R_{t,i})\phi_\varepsilon(x_{t,i}) = AAS_t.$$

**Proof.** Linearity in w with x and R preserved; see Eq. (1). □

**Lemma 3.3 Joint invariance.**

For any channel permutation $\pi$, any embedding F, and any measurement-preserving refinement $\mathcal{P}$,

$AAS_t(\pi, F, \mathcal{P}) = AAS_t.$ □

Proof. Lemma 3.1 + Lemma 3.2; sums are permutation-invariant. □

**Definition 3.3 AAS-atom.**

Let; $c_t = (c_{t,i})i.$ be contributions from Eq. (2). The multiset is AAS-atomic if the total $\sum_i c_{t,i} = AAS_t$ is invariant under the admissible class permutations, embeddings, and measurement-preserving refinements.

**Proposition 3A- Atomic character.**

Every complex structure contributes through atomic terms, and $AAS_t = \sum_i c_{t,i}$ while geometric/formal manipulations that do not alter w,R,x cannot change the total. □

**Corollary 3.1 Spatial appearance is extrinsic.**

Changing only spatial/formal metadata affects $AAS_t$ iff it changes x, R, or w; thus, space/figure are representational, not intrinsic, in this measurement.

**Monad 7 Windowlessness Causal Insulation (§7)**

**Philosophical premise.** A monad is said to have "no windows": nothing is received from without, nor are its internal states produced or steered by other created things (Leibniz, 1714/1948, §7). Within AAS, this is rendered as

(i) exclusion of external injections, (ii) cross-channel insulation at the content level, and (iii) suppression of ghost additions.

**Clause 7.1 Windowless updating.**

Channel updates are determined solely by a channel's own history $E_i(t)$, e.g., past; $\{x_{s,i}, R_{s,i}\}_{s \leq t}$, with fixed or rate-limited weights; no exogenous input $U_t$ enters. Definitions and notation follow §3.1.

**Lemma 7.1 No external accidents.**

If a hypothetical non-structural content boost $\delta_{t,i} \geq 0$ is proposed, so that the content term would read

$\phi_\varepsilon(x_{t,i}) + \delta_{t,i}$, then, because $AAS_t$ is defined only on $(w_i, R_{t,i}, x_{t,i})$ (§3.1, Eq. 1), such injections are excluded by definition and leave $AAS_t$ unchanged.

**Clause 7.2 Cross-channel insulation at content level.**

For $j \neq i$, $\frac{\partial \phi_\varepsilon(x_{t,i})}{\partial x_{t,j}} = 0$ inter-channel influence may enter the total only via the redundancy/overlap term R, which can depend on the multiset of channels, not by directly altering $\phi_\varepsilon(x_{t,i})$.

**Lemma 7.2 No induced internal motion.**

Variations in $x_{t,j}$ with $j \neq i$ can affect $c_{t,i}$ only through $R_{t,i}$; the core kernel $\phi_\varepsilon(x_{t,i})$ remains uninfluenced by other channels. Symmetrically, channel i cannot export content into $\phi_\varepsilon(x_{t,j})$ for $j \neq i$.

**Clause 7.3 Ghost suppression.**

A newly appended channel g with no internal history is assigned $w_g = 0$ or $R_{t,g} = 1$ so that $c_{t,g} = 0$; hence, ghost additions do not change $AAS_t$.

**Proposition 7A- AAS determination under windowlessness.**

Under Clauses 7.1–7.3, $AAS_t$ is fully determined by internal states and relational overlap: external injections, cross-channel content impositions, and ghost additions do not influence the score.

**Implementation note.**

In practice, append-only provenance is maintained for (x,R,w); channel firewalls, no shared mutable state or cross-reads, are enforced; and leakage audits are conducted using shuffle/drop and Granger-style tests.

**Monad 9 Identity of Indiscernibles- Intrinsic Differentiability (§9)**

Philosophical premise. Leibniz holds that "there are no two beings in nature that are perfectly identical," so individuation must rest on an intrinsic denomination rather than on relational accidents (Leibniz, 1714/1948, §9).

**Operationalization (uses §3.1).** With the core score and contributions given in Eq. (1)– (3), define for channel i at time t the intrinsic signature

$$\tau_{t,i} := (w_i, \phi_\varepsilon(x_{t,i}), \Delta\Phi_{t,i}), \text{ where } \Delta\Phi_{t,i} := \phi_\varepsilon(x_{t,i}) - \phi_\varepsilon(x_{t-1,i}).$$

The path signature is $I_i = \{\tau_{t,i}\}_{t \in \mathbb{Z}}$. Intrinsic pieces $(w_i, \phi_\varepsilon(x_{t,i}), \Delta\Phi_{t,i})$ characterize content and its transitions; the redundancy term $R_{t,i}$ is treated as extrinsic, relational.

**Axiom 9.1 intrinsic differentiation.** For any two distinct channels $i \neq j$, there exists t with $\tau_{t,i} \neq \tau_{t,j}$, equivalently, $I_i \neq I_j$. This encodes the Leibnizian requirement that individuation requires an intrinsic difference.

**Lemma 9.1 reduction under intrinsic coincidence.**

If $i \neq j$ but $\tau_{t,i} = \tau_{t,j}$ and $\alpha_{t,i} \equiv \alpha_{t,j}$ for all t, where $\alpha_{t,i} := w_i(1 - R_{t,i})$, then

$c_{t,i} = c_{t,j}$ and $AAS_t$, with i, j merged = $AAS_t$.

Consequence. Channels with identical intrinsic paths and identical scaling $\alpha_{t,\cdot}$ are deduplicated to a single representative; identity is intrinsic, not a matter of labels or routing.

**Quantifying difference.** Define the intrinsic distance

$$D(i,j) := \sup_t \left( |w_i - w_j| + |\phi_\varepsilon(x_{t,i}) - \phi_\varepsilon(x_{t,j})| + |\Delta\phi_{t,i} - \Delta\phi_{t,j}| \right).$$

Under Axiom 9.1, $D(i,j) > 0$ whenever $i \neq j$, separation; $D(i,i)=0$.

**Lemma 9.2 contribution differentiation.**

If $\alpha_{t,i} = \alpha_{t,j}$ but $x_{t,i} \neq x_{t,j}$, then

$$|c_{t,i} - c_{t,j}| = \alpha_{t,i} |\phi_\varepsilon(x_{t,i}) - \phi_\varepsilon(x_{t,j})| > 0,$$

since $\phi_\varepsilon$ is strictly decreasing and strictly convex. Intrinsic differences, not merely relational ones, therefore, yield distinct contributions.

**Lemma 9.3 limits of relational difference.**

When $\phi_\varepsilon(x_{t,i}) = \phi_\varepsilon(x_{t,j})$ and $w_i = w_j$ but $R_{t,i} \neq R_{t,j}$, the difference is purely a scaling via $\alpha_{t,i} = w_i(1 - R_{t,i})$. A change in redundancy alone is insufficient to individuate two monads.

**Conclusion.** The measurement refuses to count separate entities on the basis of relational distinctions alone. Individuation is governed by the intrinsic signature $I_i$; channels with $I_i = I_j$ are merged, whereas channels with $I_i \neq I_j$ are recognized as distinct. In this way, Leibniz's doctrine of the identity of indiscernibles is realized inside AAS as (i) a concrete signature, (ii) a distance that separates intrinsically different channels, and (iii) an explicit deduplication rule consistent with Eq. (1)– (3).

### 3.2.2 Dynamics (4)

**Monad 10 Continuous Change (§10)**

**Philosophical premise.** In §10, Leibniz holds that every created monad changes "at every moment," and that this change is continuous. In the AAS setting, this premise is taken to require smooth temporal evolution of each channel's signal.

**Clause and results (notation from §3.1; score as in Eq. (1)).**

**Continuity (Prop. 10A).** If, at a time $t_0$, both $x_i(\cdot)$ and $R_i(\cdot)$ are continuous for all i, then the per-channel terms $c_i(\cdot)$ and the total $AAS(\cdot)$ are continuous at $t_0$. The conclusion follows because $\phi_\varepsilon \circ x_i$ is continuous on $[0,1]$, products of continuous functions are continuous, and finite or uniformly bounded countable sums preserve continuity.

**Lipschitz control (Lemma 10.1 & Cor. 10A).** When $x_i, R_i$ are, absolutely, continuous with essential bounds $|x_i'|_\infty \leq L_x$ and $|R_i'|_\infty \leq L_R$,
$\frac{d}{dt}AAS(t) = \sum_i w_i \left(-R_i'(t)\, \phi_\varepsilon(x_i(t)) + (1 - R_i(t))\, \phi_\varepsilon'(x_i(t))\, x_i'(t)\right)$, with $\phi_\varepsilon'(x) = -\frac{1}{(x+\varepsilon)\ln 2}$. Hence $\left|\frac{d}{dt}AAS(t)\right| \leq \log_2\left(\frac{1+\varepsilon}{\varepsilon}\right) L_R + \frac{1}{\varepsilon \ln 2} L_x$, so $AAS(\cdot)$ is Lipschitz, in particular, piecewise $C^1$ curves are covered. This bound is used to set safe rate-limits on updates.

**No jumps (Prop. 10B).** If all $x_i(\cdot)$ and $R_i(\cdot)$ are continuous, e.g., piecewise $C^1$ without jump discontinuities, then $AAS(\cdot)$ has no jump discontinuities. Roll-outs, therefore, appear as ramps rather than spikes.

**Non-constancy under isolated variation (Lemma 10.2).** On any interval [a,b], if there exists a channel k with non-zero activity mass $\alpha_k(t) > 0$ on a set of positive measure and either $x_k$ or $R_k$ varies on a set of positive measure while all other channels are constant, then $AAS(t)$ is not constant on [a,b]. Sustained flatlines arise only through exact cross-channel cancellation of the derivative terms, which is diagnostically flagged.

**Interpretation.** The continuity and Lipschitz properties operationalize Leibniz's requirement of uninterrupted change; rate-limits and streaming estimators can be calibrated directly from the derivative bound. The "no-jumps" and "non-constancy" results separate healthy, ramp-like progress from pathological spikes or fine-tuned cancellations, enabling auditable, stable rollouts.

**Monad 11 Change from an Internal Principle (§11)**

**Philosophical premise.** In §11, Leibniz states that a monad's natural state transitions arise from an internal principle rather than from external causes. In the AAS setting, this becomes a requirement of internality: each channel's temporal evolution is driven by its own history; real-time cross-channel causation is not permitted.

**Clause and results (notation from §3.1; score as in Eq. (1)).**

**Internal transition law (Def. 11.1, INT).** For every channel i, the state pair $(x_i, R_i)$ evolves from its own local history $\mathcal{H}_i(t) = \{x_i(s), R_i(s): s \leq t\}$, and possibly a global clock t, never from $\mathcal{H}_j$ with $j \neq i$.

Discrete form: $(x_i, R_i)(t+1) = \Psi_i(\mathcal{H}_i(t), t)$.

Continuous form: $\dot{x}_i(t) = F_i(\mathcal{H}_i(t), t)$, $\dot{R}_i(t) = G_i(\mathcal{H}_i(t), t)$.

**Instantaneous isolation (Lemma 11.1).** Because $c_{t,i} = w_i(1 - R_{t,i})\phi_\varepsilon(x_{t,i})$ depends only on $(x_{t,i}, R_{t,i})$, it holds that $\partial c_{t,i}/\partial x_{t,j} = 0$ and $\partial c_{t,i}/\partial R_{t,j} = 0$ for $j \neq i$. Thus, other channels cannot directly steer the content term of i.

**Counterfactual internality (Prop. 11A).** If two worlds; $\omega$, $\omega'$ share the same $\mathcal{H}_i(t_0)$, then under INT the trajectories $(x_i, R_i)$ and contributions $c_{t,i}$ coincide for all $t \geq t_0$. The transition law for i is therefore insensitive to counterfactual changes outside $\mathcal{H}_i$.

**Separability (Cor. 11A).** If $\omega$ and $\omega'$ differ only on a subset J of channels and use the same weights, then

$$AAS^\omega(t) - AAS^{\omega'}(t) = \sum_{j \in J}(c_{t,j}^\omega - c_{t,j}^{\omega'})$$

Changes outside J cannot indirectly modify $c_{t,i}$ for $i \notin J$.

**Statistical variant (Def. 11.2).** Granger internality, discrete time: for $i \neq j$, histories of channel j do not improve prediction of $(x_i, R_i)(t+1)$ beyond $\mathcal{H}_i(t)$.

**Lemma 11.2.** INT implies Granger-internality: knowing other channels' histories confers no predictive gain for $(x_i, R_i)$.

**Derivative form and cross-channel insensitivity (Prop. 11B).** Using the derivative expression from Monad 10, first-order within-session perturbations to $x_j$ or $R_j$, $j \neq i$, leave $c_{t,i}$ unchanged under INT; effects across channels can enter the total only via the redundancy term R, not through content injection.

**Interpretation.** The internality clause realizes Leibniz's "inner source of change" by making channel updates functions of their own past. Practically, this supports auditability: shuffle/drop/Granger tests are used to check that other channels' histories add no predictive power, and simple "firewalls", no cross-reads or shared mutable state for content, enforce the clause during implementation.

**Monad 12 Particular Series of Changes (§12)**

**Philosophical premise.** Beyond the facts that change is continuous (§10) and internally driven (§11), each monad exhibits its own characteristic sequence of transitions, a particular trajectory that expresses its nature. In AAS, this becomes a channel-specific evolution law together with time-profile diagnostics.

**Clause and results (notation from §3.1; score as in Eq. (1)).**

**Trajectory law and signature (Def. 12.1).** Channel i evolves under an internal transition law discrete: $(x_i, R_i)(t+1) = \Psi_i(\mathcal{H}_i(t), t)$, continuous: $\dot{x}_i(t) = F_i(\mathcal{H}_i(t), t)$, $\dot{R}_i(t) = G_i(\mathcal{H}_i(t), t)$, possibly with parameters $\theta_i$.

The trajectory signature over horizon T is $Sig_i(T) := \{(x_{t,i}, R_{t,i})\}t = 1^T$, and its cost trace is

$c_i(T) = \{c_{t,i}\}t = 1^T$ with $c_{t,i} = w_i(1 - R_{t,i})\phi_\varepsilon(x_{t,i})$.

**Uniqueness under internality (Lemma 12.1).** If two instances of channel i share the same initial state and the same $(\Psi_i, \theta_i)$, then $\text{Sig}_i(T)$ and hence $c_i(T)$ are uniquely determined.

**Time-spread entropy (Def. 12.2; Prop. 12B).** Let $C_{i,T} = \sum_{t=1}^{T} c_{t,i} > 0$ and $p_{t|i} = c_{t,i}/C_{i,T}$.

The time-spread entropy $H_{i,T}^{(\text{time})} = -\sum_{t=1}^{T} \mathbf{1}^{T} p_{t|i} \log_2 p_{t|i}$

quantifies how concentrated the penalty is in time. It satisfies $0 \leq H_{i,T}^{(\text{time})} \leq \log_2 T$, with the lower bound for a single spike and the upper bound for perfectly even spread.

**Time-averaged state and Jensen inequality (Def. 12.3; Prop. 12A).** Set $A_{i,T} = \sum_{t=1}^{T} \alpha_{t,i}$ with $\alpha_{t,i} = w_i(1 - R_{t,i})$, and $\overline{x_{i,T}} = (\sum_t \alpha_{t,i} x_{t,i})/A_{i,T}$ (if $A_{i,T} > 0$). Since $\phi_\varepsilon$ is convex and decreasing,

$$\sum_{t=1}^{T} \alpha_{t,i} \, \phi_\varepsilon(x_{t,i}) \geq A_{i,T} \, \phi_\varepsilon(\overline{x_{i,T}}),$$

with equality iff $x_{t,i}$ is constant in t. Thus, holding $A_{i,T}$ and $\overline{x_{i,T}}$ fixed, greater temporal heterogeneity, incurs a larger cumulative penalty, Leibniz's "particular series" becomes measurable.

**Inter-trajectory distinguishability (Def. 12.4; Lemma 12.2).** For channels i,j and horizon T, define the trajectory distance

$D_T(i,j) = \sum_{t=1}^{T} |c_{t,i} - c_{t,j}|$. Then $D_T(i,j) = 0$ iff $c_{t,i} \equiv c_{t,j}$ for all t, AAS-observational equivalence; if $D_T(i,j) > 0$, the trajectories are separable.

**Interpretation.** Monad 12 turns the qualitative idea of a "particular series of changes" into quantitative summaries, $C_{i,T}, \overline{x_{i,T}}, H_{i,T}^{(\text{time})}, D_T$ that reveal whether a channel's evolution is steady or volatile, and whether two channels follow measurably distinct courses. Combined with §10–§11, this yields an auditable, trajectory-aware account of how internally driven change accumulates in the AAS.

### Monad 15 Internal Transition Activity: Appetition (§15)

**Philosophical premise.** In §15, Leibniz characterizes appetition as the inner activity that carries a monad from one perception to the next. Access to the intended percept may be partial, but directed advancement is secured and culminates in a new perceptual state (Leibniz, 1714/1948, §15).

**Clause and results (notation from §3.1; score as in Eq. (1)).**

**Internal target and step.** For channel i at time t, appetition specifies an internal goal $g_{t,i} \in [0,1]$ and a step coefficient $\eta_{t,i} \in [0,1]$. The state update is

$x_{t+1,i} = (1 - \eta_{t,i}) x_{t,i} + \eta_{t,i} g_{t,i}.$

Here $\eta_{t,i} = 0$ means "halt," $\eta_{t,i} = 1$ "full access," and $0 < \eta_{t,i} < 1$ "partial access."

**Lemma 15.1 partial access ⇒ quantified gap closure.** Because $\phi_\varepsilon$ is decreasing and convex,

$$\phi_\varepsilon(x_{t,i}) - \phi_\varepsilon(x_{t+1,i}) \geq \eta_{t,i}\left(\phi_\varepsilon(x_{t,i}) - \phi_\varepsilon(g_{t,i})\right).$$

Thus, if the target is penalty-improving $\phi_\varepsilon(g_{t,i}) \leq \phi_\varepsilon(x_{t,i})$, the penalty gap shrinks by at least the factor $\eta_{t,i}$.

**Definition 15.2 new-perception magnitude.**

$$N^t := \sum_i w_i(1 - R_{t,i})|x_{t+1,i} - x_{t,i}| = \sum_i w_i(1 - R_{t,i})\eta_{t,i}|g_{t,i} - x_{t,i}|.$$

If $N^t > 0$, a new percept has occurred, some channel changed with nonzero weighted step.

**Proposition 15A- AAS decreases under benevolent appetition.**

If $\phi_\varepsilon(g_{t,i}) \leq \phi_\varepsilon(x_{t,i})$ for all i, then the score in Eq. (1) satisfies

$$AAS_t - AAS_{t+1} \geq \sum_i w_i(1 - R_{t,i})\eta_{t,i}\left(\phi_\varepsilon(x_{t,i}) - \phi_\varepsilon(g_{t,i})\right) \geq 0,$$

with equality only when $\eta_{t,i} = 0$ or $g_{t,i} = x_{t,i}$. Hence, even partial, benevolent appetition guarantees non-negative progress and reduces the overall penalty.

**Bound 15.1 (small-step regularity).** Since $\phi'_\varepsilon(x) \leq -[(x+\varepsilon)\ln 2]^{-1}$,

$$|\phi_\varepsilon(x_{t+1,i}) - \phi_\varepsilon(x_{t,i})| \leq \frac{1}{\varepsilon \ln 2}|x_{t+1,i} - x_{t,i}| = \frac{\eta_{t,i}}{\varepsilon \ln 2}|g_{t,i} - x_{t,i}|.$$

Appetitive steps are therefore Lipschitz-controlled; no jumps or pathologies arise when steps are rate-limited.

**Interpretation.** Appetition functions as an internal controller: targets that improve the penalty drive AAS downward, while the step coefficient $\eta$ provides a quantitative knob for safe, smooth advancement. In practice, logging; $g_{t,i}, \eta_{t,i}$, $N^t$ enables auditable progress, and rate-limiting $\eta$ enforces stability consistent with the continuous-change theme of §10–§12.

### 3.2.3 Representation&Consciousness (4)

**Monad 14 Perception as Unified Multiplicity; Perception ≠ Apperception (§14)**

**Philosophical premise.** In §14, Leibniz characterizes perception as a transient state that represents multiplicity within unity; it is distinct from apperception, self-conscious awareness. Conflating the two leads to the mistaken view that sustained unconsciousness is equivalent to death (Leibniz, 1714/1948, §14).

**Notation.** All symbols are as defined in 3.1. The only new symbol here is the number of active channels

$$m_t := |\{i: p_{t,i} > 0\}|.$$

**Perception.** Perception is present iff $S_t > 0$.

**Apperception (distinctiveness).** Apperceptive intensity rises as the peak share $\rho_t$ increases and the spread/suppression $\kappa_t$ decreases. The salience scale $\text{AppearLevel}_t = (1 - \kappa_t)\,\rho_t$ (from 3.1) captures this continuously and without thresholds.

**Definition 14.1 Perception.** There is perception at time t iff the total contribution $S_t$ is strictly positive. Unity is captured by the scalar $S_t$; multiplicity is captured by the distribution $\{p_{t,i}\}i: p_{t,i} > 0$.

**Definition 14.2 Apperceptive distinctiveness.** For $m_t \geq 2$, distinctiveness at time t is summarized by the pair $(\rho_t, \kappa_t)$, where $\rho_t \in [1/m_t, 1]$ quantifies dominance and $\kappa_t \in [0,1]$ quantifies normalized spread.

**Lemma 14.1 Multiplicity–distinctiveness tension.**

$$0 \leq H_t^{(\text{contrib})} \leq \log_2 m_t.$$

Extremes: $H_t^{(\text{contrib})} = 0 \iff \rho_t = 1$ (single-channel dominance);

$H_t^{(\text{contrib})} = \log_2 m_t \iff p_{t,i} = 1/m_t$ (uniform spread).

**Lemma 14.2 Peak–entropy bound.**

For fixed $m_t \geq 2$ and peak $\rho_t$,

$$H_t^{(\text{contrib})} \leq -\rho_t \log_2 \rho_t - (1-\rho_t)\log_2\left(\frac{1-\rho_t}{m_t - 1}\right),$$

with equality for the "single peak + uniform background" distribution. As $\rho_t$ increases, the bound decreases, quantifying the emergence of a salient figure from background.

**Proposition 14.1 Bounds on salience.**

Since $\rho_t \in [1/m_t, 1]$ and $\kappa_t \in [0,1]$, the salience scale satisfies

$$\text{AppearLevel}_t = (1 - \kappa_t)\,\rho_t \in \left[\frac{1-\kappa_t}{m_t}, 1 - \kappa_t\right].$$

Thus $\text{AppearLevel}_t$ increases monotonically when the peak strengthens or the spread/suppression weakens.

### 14.3. Reply to the Cartesian Objection

**Proposition 14A Unconscious perceptions exist.**

It may occur that $S_t > 0$ while $\rho_t \approx 1/m_t$ and $\kappa_t$ is high: there is non-zero perceptual mass, many tiny contributions, but no dominant share. This provides a precise counterexample to "no perception without consciousness."

**Corollary 14A Prolonged unconsciousness $\neq$ death.**

A system can maintain $S_t > 0$ for extended periods with diffuse perception, low $\rho_t$, high $\kappa_t$. True cessation is marked by $S_t = 0$, all contributions vanish, not by a low-but-positive $S_t$.

**Monad 21 No Perceptionless State; Multiplicity Yields Dizziness (§21)**

**Philosophical premise.** Even in fainting or deep sleep, a simple substance is never without perception. What disappears is distinct awareness: when there are many extremely small perceptions, none is salient, producing a blur or dizziness (Leibniz, 1714/1948, §21).

**Notation.** All symbols follow 3.1. New here: a saliency threshold, $\tau > 0$; the number of active channels $m_t := |\{i: c_{t,i} > 0\}|$; the salient mass $S_t^{(\tau)} := \sum_i c_{t,i} \mathbf{1}\{c_{t,i} \geq \tau\}$; the positive–part operator $[z]_+ := \max\{z, 0\}$.

### 21.1. Formalization in AAS

**Definition 21.1 $\tau$-dizziness.**
At time t, the system is in $\tau$-dizziness if $AAS_t > 0$ and $S_t^{(\tau)} = 0$, equivalently, $\max_i c_{t,i} < \tau$. Then the entire perceptual mass consists of sub-threshold micro-signals: multiplicity without a distinguishable figure.

**Proposition 21.1 No perceptionless state ⇒ AAS positivity.**
If some channel i satisfies $w_i > 0$, $x_{t,i} < 1$ and $R_{t,i} < 1$ then $AAS_t = \sum_i c_{t,i} > 0$.

Proof sketch. By 3.1, $\phi_\varepsilon(1) = 0$ and $\phi_\varepsilon$ is strictly decreasing, so $x_{t,i} < 1 \Rightarrow \phi_\varepsilon(x_{t,i}) > 0$.

With $w_i > 0$ and $1 - R_{t,i} > 0$, $c_{t,i} > 0$, is obtained, hence $S_t > 0$ and $AAS_t > 0$. □

**Corollary 21.1 Multiplicity cost of dizziness.**
Under $\tau$-dizziness,
$$AAS_t = \sum_i c_{t,i} \leq m_t \tau \quad \Rightarrow \quad m_t \geq \frac{AAS_t}{\tau}.$$
Thus, in the absence of any salient contribution, a positive $AAS_t$ requires many micro-perceptions, quantitatively capturing Leibniz's "dizziness from multiplicity."

### 21.2. Apperceptive Dizziness via Lack of Sharp Change

Define the one-step penalty rise

$$\Delta\Phi_{t,i} := [\phi_\varepsilon(x_{t,i}) - \phi_\varepsilon(x_{t-1,i})]_+, \quad AP_t := \max_i \Delta\Phi_{t,i}.$$

**Definition 21.2 $\delta$-dizziness.**
For a sharpness threshold $\delta > 0$, declare $\delta$-dizziness when $AAS_t > 0$ and $AP_t < \delta$. In this regime, the system carries non-zero perceptual mass, yet no channel exhibits a sharp local jump; consequently, no figure emerges from background, and apperception remains weak.

**Interpretation.**
Leibniz's claim in §21 is realized in two complementary ways: (i) mass without salience ($\tau$-dizziness) and (ii) change without edge ($\delta$-dizziness). Both maintain $AAS_t > 0$, perception, persists, while keeping apperceptive distinctiveness low no dominant share or sharp rise. This distinguishes prolonged unconsciousness from true cessation $S_t = 0$.

**Monad 26 Memory Gives Sequence, Resembling Reason but Not Being Reason (§26)**

**Philosophical premise.** Memory sustains sequence by retaining associations from prior, similar perceptions, supporting expectation and behavioral continuity. Yet memory is not reason: reason proceeds from necessary truths, whereas memory tracks resemblance, recurrence, and association (Leibniz, 1714/1948, §26).

**Notation.** All symbols follow 3.1. New here:

forgetting parameter $\lambda \in (0,1)$;

one-step penalty rise $\Delta\Phi_{t,i} := [\phi_\varepsilon(x_{t,i}) - \phi_\varepsilon(x_{t-1,i})]_+$;

memory trace $M_{t,i}^{(\lambda)} := \sum_{k=1}^{\infty} \lambda^{k-1} \Delta\Phi_{t-k,i} \geq 0$;

total trace mass $\|M_t^{(\lambda)}\|_1 := M_{t,j}^{(\lambda)}$ (if $> 0$);

normalized sequential prior $q_{t,i}^{(\lambda)} := \dfrac{M_{t,i}^{(\lambda)}}{\|M_t^{(\lambda)}\|_1}$, when $\|M_t^{(\lambda)}\|_1 > 0$

## 26.1. Memory Trace and Sequential Distribution

**Definition 26.1 Sequential prior.**

$q_t^{(\lambda)}$ encodes a residual memory of past saliences. Expectation means the present penalty distribution $p_t$ (from 3.1) tracks $q_t^{(\lambda)}$.

**Definition 26.2 Sequentiality measures.**

$$\mathrm{Consec}_t^{(\lambda)} := \sum_{i:\, M_{t,i}^{(\lambda)}>0} q_{t,i}^{(\lambda)}\, p_{t,i} \in [0,1], \quad D_t^{(\lambda)} := \sum_i p_{t,i} \log_2 \frac{p_{t,i}}{q_{t,i}^{(\lambda)}} \, (\geq 0).$$

If $\operatorname{supp} p_t \not\subseteq \operatorname{supp} q_t^{(\lambda)}$, compute $D_t^{(\lambda)}$ with a smoothed prior $\hat{q}$.

**Lemma 26.1 Bounds.**

$0 \leq \mathrm{Consec}_t^{(\lambda)} \leq 1$ and $D_t^{(\lambda)} \geq 0$, with $\mathrm{Consec}_t^{(\lambda)}=1$ and $D_t^{(\lambda)}=0$ iff $p_t \equiv q_t^{(\lambda)}$.

Interpretation. Memory acts as sequential expectation: when a stimulus recurs, the channel previously associated with salience has large $M_{t,i}^{(\lambda)}$; if current penalty concentrates there, $\mathrm{Consec}_t^{(\lambda)}$ is high and $D_t^{(\lambda)}$ low.

## 26.2. Salient Events and Expectation Mass

Fix a salience threshold $\tau > 0$.

**Definition 26.3 Salient event; expectation mass.**

A salient event occurs on channel i when $\Delta\Phi_{t,i} \geq \tau$. Define the expectation mass

$$B_t^{(\lambda)}(\tau) := \sum_{i: M_{t,i}^{(\lambda)} \geq \tau} c_{t,i}.$$

**Proposition 26A Lower bound on consecutiveness.**

If $S_t > 0$ and $\|M_t^{(\lambda)}\|_1 > 0$, $\quad\quad \text{Consec}_t^{(\lambda)} \geq \dfrac{\tau}{\|M_t^{(\lambda)}\|_1} \times \dfrac{B_t^{(\lambda)}(\tau)}{S_t} \leq 1.$

Proof sketch. Since $q_{t,i}^{(\lambda)} = M_{t,i}^{(\lambda)}/\|M_t^{(\lambda)}\|_1$, the set $\{i: M_{t,i}^{(\lambda)} \geq \tau\}$ equals $\{i: q_{t,i}^{(\lambda)} \geq \frac{\tau}{\|M_t^{(\lambda)}\|_1}\}$. Then

$$\text{Consec}_t^{(\lambda)} = \sum_i q_{t,i}^{(\lambda)} p_{t,i} \geq \sum_{i: M_{t,i}^{(\lambda)} \geq \tau} q_{t,i}^{(\lambda)} p_{t,i} \geq \frac{\tau}{\|M_t^{(\lambda)}\|_1} \sum_{i: M_{t,i}^{(\lambda)} \geq \tau} p_{t,i} = \frac{\tau}{\|M_t^{(\lambda)}\|_1} \times \frac{B_t^{(\lambda)}(\tau)}{S_t}$$

The upper bound $\leq 1$ is immediate from the definition of $\text{Consec}_t^{(\lambda)}$. □

**Lemma 26.2 No memory overflow.**

Since $0 \leq \Delta\Phi_{t,i} \leq \log_2 \frac{1+\varepsilon}{\varepsilon}$, $M_{t,i}^{(\lambda)} \leq \frac{1}{1-\lambda} \phi_\varepsilon(0)$, so $\|M_t^{(\lambda)}\|_1 < \infty$

for all t. Thus, the sequential prior remains well-defined, no pathological growth.

### 26.3. Memory Is Not Reason

**Result 26.1 Resemblance vs. necessity.**

High $\text{Consec}_t^{(\lambda)}$ with low $D_t^{(\lambda)}$ indicates alignment with past salience patterns, that is, resemblance, recurrence, and association, not derivation from necessary truths. Memory, therefore resembles reason in guiding expectation but is not reason. This sets up the contrast with the treatment of necessary and eternal truths in §29.

**Monad 29 Necessary/Eternal Truths = Reason (ratio) (§29)**

**Philosophical premise.** Leibniz separates animal memory from reason by appeal to necessary and eternal truths. Memory leans on similarity and repetition (§26); reason carries law-like necessity and justification (Leibniz, 1714/1948, §29).

**Notation.** Symbols follow 3.1. Additionally, is used:

$q_t^{(\lambda)}$: the sequential habit prior from §26 (e.g., an EMA of $p_t$);

$r \in \Delta^m$: a time-invariant rational prior encoding necessary truths;

support assumption for finite KL: after standard η-smoothing, $r_i > 0$ and $q_{t,i}^{(\lambda)} > 0$ for all i.

### 29.1. Evidence for Reason

**Definition 29.1 Reason score.**

$$\text{Reason}_t := D(p_t \| q_t^{(\lambda)}) - D(p_t \| r) = \sum_i p_{t,i} \log_2 \frac{r_i}{q_{t,i}^{(\lambda)}}.$$

**Lemma 29.1 Sign.**

$\text{Reason}_t \geq 0$ iff $p_t$ is closer to $r$ than to $q_t^{(\lambda)}$ in KL. In particular,

if $p_t = r$, then $\text{Reason}_t = D(r \| q_t^{(\lambda)}) \geq 0$;

If $p_t = q_t^{(\lambda)}$, then $\text{Reason}_t = -D(q_t^{(\lambda)} \| r) \leq 0$.

**Reading.** Increasing $\text{Reason}_t$ formalizes Leibniz's claim that reason "elevates" us beyond habit: the instantaneous structure aligns more with a law-like prior than with historical traces.

### 29.2. Truth Floors and AAS Bounds

Let $S \subseteq A_t$ collect channels that encode necessary/eternal truths, and fix a truth floor $\beta \in [0,1]$.

**Definition 29.2 Truth threshold.**

In the rational regime, a uniform floor is imposed: $\forall i \in S,\ x_{t,i} \geq \beta$.

**Proposition 29A Guaranteed upper bound from reason.**

For any t,

$$\text{AAS}_t = \sum_{i \in S} \alpha_{t,i}\, \phi_\varepsilon(x_{t,i}) + \sum_{i \notin S} \alpha_{t,i}\, \phi_\varepsilon(x_{t,i}) \leq \left(\sum_{i \in S} \alpha_{t,i}\right) \phi_\varepsilon(\beta) + \sum_{i \notin S} \alpha_{t,i}\, \phi_\varepsilon(x_{t,i}).$$

Proof. $\phi_\varepsilon$ is decreasing; thus $x_{t,i} \geq \beta \Rightarrow \phi_\varepsilon(x_{t,i}) \leq \phi_\varepsilon(\beta)$ on S. □

**Meaning.** Let $\alpha_{S,t} := \sum_{i \in S} \alpha_{t,i}$. Larger $\alpha_{S,t}$ and higher $\beta$ tighten a time-independent cap on $\text{AAS}_t$. Unlike memory, which depends on history, reason secures an a priori bound via necessary truths.

**Result 29.1 Two global upper bounds.**

(a) $\alpha$-mass bound. Define the $\alpha$-share on rational channels

$$\rho_t^{(\alpha)} := \frac{\sum_{i \in S} \alpha_{t,i}}{A_t}, \qquad \phi_\varepsilon(0^+) := \lim_{x \to 0^+} \phi_\varepsilon(x) = \lim_{x \to 0^+} \log_2\left(\frac{1+\varepsilon}{x+\varepsilon}\right) = \log_2\left(\frac{1+\varepsilon}{\varepsilon}\right).$$

Then

$$\text{AAS}_t = S_t \leq A_t\big[\rho_t^{(\alpha)}\phi_\varepsilon(\beta) + \big(1-\rho_t^{(\alpha)}\big)\phi_\varepsilon(0^+)\big].$$

Derivation. Apply Proposition 29A on S and the trivial $\phi_\varepsilon(x) \leq \phi_\varepsilon(0^+)$ on $S^c$.

(b) p-mass (instantaneous share) bound. Let

$$\rho_t := \sum_{i \in S} p_{t,i} = \frac{\sum_{i \in S} \alpha_{t,i}\,\phi_\varepsilon(x_{t,i})}{S_t}.$$

If $\rho_t > 0$, then

$$S_t \leq \frac{\phi_\varepsilon(\beta)}{\rho_t}\sum_{i \in S}\alpha_{t,i}.$$

Derivation. From Proposition 29A, $\sum_{i \in S}\alpha_{t,i}\,\phi_\varepsilon(x_{t,i}) \leq \phi_\varepsilon(\beta)\sum_{i \in S}\alpha_{t,i}$; the left side equals $\rho_t S_t$.

**Remark Do not conflate averages.**

The $\alpha$-weighted mean $\overline{\phi}_t^{(\alpha)} := S_t/A_t$ differs from the p-weighted mean

$$\overline{\phi}_t^{(p)} := \sum_i p_{t,i}\,\phi_\varepsilon(x_{t,i}) = \frac{1}{S_t}\sum_i \alpha_{t,i}\,[\phi_\varepsilon(x_{t,i})]^2.$$

They serve distinct roles and should not be interchanged.

### 29.3. Scientific Closure / Law-Fixity over Time

To capture the "eternal", law-like, time-invariant, aspect of reason, define for a horizon $T \geq 2$:

$$\text{LawFixity}_T := 1 - \frac{1}{T-1}\sum_{t=2}^{T} \text{TV}(r,\widehat{r}_t) \in [0,1],$$

where $\widehat{r}_t$ is the law inferred from $p_t$, and TV is the total-variation distance. $\text{LawFixity}_T = 1$ when the inferred law is perfectly time-invariant and aligned with necessary truths. Convention: take $\text{LawFixity}_1$ as undefined or $=1$ by fiat.

### 29.4. Synthesis: Reason vs. Memory

$\text{Reason}_t = D\big(p_t\|q_t^{(\lambda)}\big) - D(p_t\|r)$ measures instantaneous preference for law over habit; non-negativity indicates rational alignment. Truth floors, $x_{t,i} \geq \beta$ on S, upper-bound the penalty: Proposition 29A provides a pointwise cap, while Result 29.1 offers global caps via $\alpha$- and p-shares. High $\text{LawFixity}_T$ formalizes the Leibnizian idea that reason grasps time-invariant principles; memory alone (cf. §26) cannot deliver this fixity.

### 3.2.4 Coherence and Rational Necessity (4)

**Monad 31 Principle of Contradiction (PC) (§31)**

**Philosophical premise.** Leibniz grounds rational inference on (i) the Principle of Contradiction, any proposition containing an internal contradiction is false, and (ii) the Principle of the Excluded Middle (Leibniz, 1714/1948, §31). Here, PC is formalized within the AAS architecture: simultaneous affirmation of logically contradictory contents is mathematically prohibited and incurs a quantifiable penalty.

**Notation.** All symbols follow 3.1. New in this subsection:

$\mathcal{E} \subseteq \{(i,j): i \neq j\}$: set of mutually exclusive, logically contradictory channel pairs.

$\gamma_{i,j} \geq 0$: penalty weight for pair $(i,j) \in \mathcal{E}$.

$\zeta \in [0,1]$: selectivity margin, treats "effectively zero" truth up to $\zeta$.

$m_{ij} := \min\{x_{t,i}, x_{t,j}\}, \qquad m_{ij}^{(\zeta)} := \max\{0, m_{ij} - \zeta\}.$

**PC penalty**

$$PC_t = \sum_{(i,j) \in \mathcal{E}} \gamma_{i,j} \, \phi_\varepsilon\left(1 - m_{ij}^{(\zeta)}\right).$$

Intuition. If both members of a contradictory pair carry high truth ($x_{t,i}, x_{t,j}$ large), then $m_{ij}^{(\zeta)}$ rises, the argument $1 - m_{ij}^{(\zeta)}$ falls, and, since $\phi_\varepsilon$ is strictly decreasing, the penalty increases. If at least one side is effectively null ($m_{ij} \leq \zeta$), then $m_{ij}^{(\zeta)} = 0$ and the added penalty is zero.

**31.1. Basic Properties**

**Lemma 31.1 Monotonicity, zero condition, uniform bound.** For any $(i,j) \in \mathcal{E}$:

**Monotonicity.** $m_{ij} \mapsto \phi_\varepsilon\left(1 - m_{ij}^{(\zeta)}\right)$ is strictly increasing.

**Zero with margin.** $\phi_\varepsilon\left(1 - m_{ij}^{(\zeta)}\right) = 0$ iff $m_{ij} \leq \zeta$.

**Uniform upper bound.** $\phi_\varepsilon\left(1 - m_{ij}^{(\zeta)}\right) \leq \phi_\varepsilon(0^+) = \log_2 \frac{1+\varepsilon}{\varepsilon}$.

Proof. $m \mapsto (m - \zeta)^+$ is increasing; $\phi_\varepsilon$ is decreasing with $\phi_\varepsilon(1) = 0$; the upper bound follows from

$$1 - m_{ij}^{(\zeta)} \in (0,1]. \square$$

**Lemma 31.2 Symmetry maximizes PC under fixed total truth.**

Let $s_{ij} := x_{t,i} + x_{t,j}$ be fixed. With $\zeta = 0$,

$$\phi_\varepsilon(1 - \min\{x_{t,i}, x_{t,j}\}) \leq \phi_\varepsilon\left(1 - \frac{s_{ij}}{2}\right),$$

with equality iff $x_{t,i} = x_{t,j} = \frac{s_{ij}}{2}$. Thus, splitting truth symmetrically across a contradictory pair maximizes the PC penalty; placing "all truth on one side" minimizes it. The same conclusion holds for $\zeta > 0$ whenever $s_{ij} > 2\zeta$, with min replaced by $m_{ij}^{(\zeta)}$.

### 31.2. Selectivity and Excluded Middle (Operational Form)

**Proposition 31A PC-favorable distributions via local selectivity.**

If for every $(i,j) \in \mathcal{E}$, $m_{ij} \leq \zeta$, then $PC_t = 0$. Conversely, if some pair satisfies $m_{ij} > \zeta$, then $PC_t > 0$.

**Operational reading (Excluded Middle).**

To minimize $AAS_t^{(31)}$, each contradictory pair must receive exclusive truth assignment, up to the margin $\zeta$. In words: do not make both sides true.

**Corollary 31.1 Contradiction-free configurations minimize the total.**

Fix $\Gamma := \sum_{(i,j) \in \mathcal{E}} \gamma_{i,j}$. Over the feasible hypercube $(0,1]^m$, the minimum of $AAS_t^{(31)}$ is attained when $m_{ij} \leq \zeta$ for all $(i,j) \in \mathcal{E}$ hence $PC_t = 0$; remaining channels then follow the convex lower envelope of the base $AAS_t$.

### 31.3. Global Bounds and Calibration
**Uniform bound on PC.**

$$0 \leq PC_t \leq \Gamma \phi_\varepsilon(0^+) = \Gamma \log_2 \frac{1+\varepsilon}{\varepsilon}.$$

**Total with PC.**

$AAS_t^{(31)} = AAS_t + PC_t \leq AAS_t + \Gamma \log_2 \frac{1+\varepsilon}{\varepsilon}$.

Calibration. The pair-weights $\gamma_{i,j}$ set how costly contradictions are; the margin $\zeta$ sets how strictly exclusivity is enforced smaller $\zeta \Rightarrow$ stricter.

### 31.4. Conclusion

By adding a contradiction penalty $PC_t$ to the AAS framework, logical incoherence receives an explicit, tunable cost. Symmetric dual assignment across a contradictory pair maximizes the penalty, selective assignment eliminates it. Thus Leibniz's Principle of Contradiction in §31 becomes a precise, operational constraint on representation that systematically enforces logical coherence within the proposed AAS architecture.

**Monad 32 Principle of Sufficient Reason (PSR): "No truth without a reason" (§32)**

**Philosophical premise.** Leibniz's PSR states that no fact or truth exists without a sufficient reason. Reasons may be opaque to us, yet they must exist. Within AAS, PSR appears as a penalty that quantifies the gap between a claim and its justificatory support.

**Notation.** All symbols follow 3.1. New here: a directed causal network G over channels with edge weights $a_{ij} \geq 0$, influence of $j \to i$, and a self-inertia weight $a_{i0} \geq 0$. Impose the budget

$$\sum_j a_{ij} + a_{i0} \leq 1.$$

so causal support is bounded.

**32.1. Causal mass, sufficiency ratio, PSR penalty**

**(i) Causal mass.** For channel i at time t,

$$r_{t,i} := a_{i0} x_{t-1,i} + \sum_j a_{ij} x_{t,j} \in [0,1],$$

interpreted as the weighted mass of reasons supporting the claim on i.

**(ii) Sufficiency ratio.** With a small $\delta > 0$ for numerical stability,

$$s_{t,i} := \min\left\{1, \frac{r_{t,i} + \delta}{x_{t,i} + \delta}\right\} \in [0,1],$$

This symmetric smoothing yields $s_{t,i}=1$ whenever reasons are sufficient $r_{t,i} \geq x_{t,i}$, ensuring zero PSR penalty and avoiding division by zero when $x_{t,i}$ is very small.

**(iii) PSR penalty and adjusted score.**

$$PSR_t \equiv \sum_i \alpha_{t,i} \, \phi_\varepsilon(s_{t,i}), \qquad AAS_t^{(32)} := AAS_t + PSR_t.$$

**32.2. Basic properties**

**Lemma 32.1 Zero, monotonicities, bound.** For each i:

**a. Zero condition.** If $r_{t,i} \geq x_{t,i}$, then $s_{t,i} = 1$ and $\phi_\varepsilon(s_{t,i}) = 0$.

**b. Monotonicity in reasons.** With $x_{t,i}$ fixed, increasing $r_{t,i}$ increases $s_{t,i}$ and decreases the penalty (since $\phi_\varepsilon$ is decreasing).

**c. Monotonicity in claim.** With $r_{t,i}$ fixed, increasing $x_{t,i}$ decreases $s_{t,i}$ and increases the penalty.

**d. Uniform bound.** Because $s_{t,i} \in [0,1]$,

$$\text{PSR}_t \leq \left(\sum_i \alpha_{t,i}\right)\phi_\varepsilon(0) = A_t \log_2 \frac{1+\varepsilon}{\varepsilon}$$

Interpretation. A claim that outstrips its reasons is penalized; sufficient reasons switch the PSR term off.

### 32.3. Optimality and extremes

**Proposition 32A PSR-compliant configurations are optimal.**

For fixed $\alpha, \varepsilon$, and network G, $\text{AAS}_t^{(32)}$ is minimized whenever $x_{t,i} \leq r_{t,i}$ for all i; in this case $\text{PSR}_t = 0$.

Proof sketch. Each summand $\alpha_{t,i}\, \phi_\varepsilon(s_{t,i})$ is minimized at $s_{t,i}=1$. Enforcing $r_{t,i} \geq x_{t,i}$ pointwise gives $s_{t,i} = 1$ and drives $\text{PSR}_t$ hence $\text{AAS}_t^{(32)}$ to its minimum. □

**Corollary 32.1 No reason ⇒ maximal penalty.**

If $r_{t,i} = 0$ while $x_{t,i} > 0$, then

$$s_{t,i} = \delta/(x_{t,i} + \delta) \in (0,1), \quad \phi_\varepsilon(s_{t,i}) < \phi_\varepsilon(0),$$

and as $\delta \downarrow 0$ the term approaches $\phi_\varepsilon(0) = \log_2 \frac{1+\varepsilon}{\varepsilon}$.

An unsupported claim is penalized as harshly as the model allows.

**Proposition 32B Unjustified inflation strictly increases the penalty.**

Fix the causal masses $\{r_{t,i}\}_i$. If some claims $\{x_{t,i}\}_i$ are inflated beyond their causal support ($x_{t,i} > r_{t,i}$), then $\text{PSR}_t$ increases strictly.

Reason. On the active region $s_{t,i} = (r_{t,i} + \delta)/(x_{t,i} + \delta) < 1$, one has $\partial s_{t,i}/ \partial x_{t,i} < 0$, hence $\partial\, \phi_\varepsilon(s_{t,i})/ \partial x_{t,i} > 0$. Summing with $\alpha_{t,i} \geq 0$ yields a strict increase whenever at least one index lies in the active region. □

### 32.4. Conclusion

PSR becomes an explicit, testable constraint within AAS: reasons at least as large as claims yield zero penalty, and any deficit is charged according to a convex, logarithmic scale. This operationalizes Leibniz's dictum, no truth without a reason, as a regularizer that favors justificatory adequacy over unsupported assertion.

### Monad 78 Pre-Established Harmony: Separate Laws, Coordinated Flow (§78)

**Philosophical premise.** Leibniz holds that soul and body follow their own laws and do not causally interact; yet, by pre-established harmony, they express one and the same world in a coordinated manner (Leibniz, 1714/1948, §78). The AAS framework gives a precise, testable formalization.

**Notation.** Symbols follow 3.1. New here: two disjoint channel sets I (soul-view) and $\mathcal{J}$ (body-view); view-specific scores $x_{t,i}^{(S)}$ (i ∈ I ) and $x_{t,j}^{(B)}$ (j ∈ $\mathcal{J}$); view-specific weights/redundancies and hence $\alpha_{t,i}^{(S)}, \alpha_{t,j}^{(B)}$ defined analogously to 3.1; a pairing map h: $\mathcal{J}$ → I matching each body channel j to a soul channel h(j).

### 78.1 Two-view AAS and the harmony penalty

Each view computes its own AAS independently:

$$\text{AAS}_t^{(S)} = \sum_{i \in I} \alpha_{t,i}^{(S)} \phi_\varepsilon(x_{t,i}^{(S)}), \qquad \text{AAS}_t^{(B)} = \sum_{j \in \mathcal{J}} \alpha_{t,j}^{(B)} \phi_\varepsilon(x_{t,j}^{(B)}),$$

Alignment between paired channels is measured by the harmony score

$$m_{t,j} := 1 - \left| x_{t,h(j)}^{(S)} - x_{t,j}^{(B)} \right| \in [0,1],$$

and penalized by

$$\text{HARM}_t := \sum_{j \in \mathcal{J}} \beta_{t,j} \phi_\varepsilon(m_{t,j}), \qquad \beta_{t,j} := \min\left\{\alpha_{t,h(j)}^{(S)}, \alpha_{t,j}^{(B)}\right\}.$$

The joint metric that incorporates both views and their alignment is

$$\text{AAS}_t^{(78)} = \text{AAS}_t^{(S)} + \text{AAS}_t^{(B)} + \text{HARM}_t$$

**Interpretation.** In line with "windowlessness," there are no causal cross-terms from soul to body or body to soul in AAS; coupling occurs only through the alignment term. When paired channels agree exactly, the penalty vanishes.

### 78.2 Properties
**Lemma 78.1 Zero-penalty condition.**

If $x_{t,h(j)}^{(S)} = x_{t,j}^{(B)}$ for all j ∈ $\mathcal{J}$, then $m_{t,j} = 1$ and $\text{HARM}_t = 0$. Hence $\text{AAS}_t^{(78)} = \text{AAS}_t^{(S)} + \text{AAS}_t^{(B)}$.

**Lemma 78.2 Monotonicity and local sensitivity.**

(i) As the mismatch $|x_{t,h(j)}^{(S)} - x_{t,j}^{(B)}|$ increases, $m_{t,j}$ decreases and $\phi_\varepsilon(m_{t,j})$ increases.

(ii) Larger $\beta_{t,j}$ make a channel contribute proportionally more to $\text{HARM}_t$.

(iii) For a small discrepancy $m_{t,j} = 1 - \delta$ with $\delta \ll 1$,

$$\phi_\varepsilon(m_{t,j}) = \log_2 \frac{1 + \varepsilon}{1 - \delta + \varepsilon} \approx \frac{\delta}{(1 + \varepsilon) \ln 2}$$

so the penalty grows linearly in the size of small misalignments.

### 78.3 Optimization and bounds

**Proposition 78A Harmony optimization.**

For fixed $\alpha^{(S)}, \alpha^{(B)}, \beta, \varepsilon$, the total $AAS_t^{(78)}$ is minimized when $x_{t,h(j)}^{(S)} = x_{t,j}^{(B)}$ for all $j \in \mathcal{J}$, i.e., $m_{t,j} = 1$ and $HARM_t = 0$.

Reason: $\phi_\varepsilon$ is convex with minimum at x=1.

**Global bound.** Since $m_{t,j} \in [0,1]$,

$$0 \leq HARM_t \leq \left(\sum_{j \in \mathcal{J}} \beta_{t,j}\right) \phi_\varepsilon(0) = \left(\sum_{j \in \mathcal{J}} \beta_{t,j}\right) \log_2 \frac{1+\varepsilon}{\varepsilon}.$$

### 78.4 Modeling principle: a pre-established source

To explain coordination without causal linkage, posit a shared latent driver $z_t$ and view-specific laws $F_S$ and $F_B$:

$$x_t^{(S)} := F_S(z_t), \quad x_t^{(B)} := F_B(z_t).$$

If the alignment constraint holds for all j and all z, $F_{S,h(j)}(z) \equiv F_{B,j}(z)$,

then $x_{t,h(j)}^{(S)} = x_{t,j}^{(B)}$ and thus $m_{t,j} \equiv 1$. Each view keeps its own internal law $F_S$, $F_B$, yet both express a coordinated flow via the common latent source, an operational rendering of Leibniz's pre-established harmony.

### Monad 79 Alignment of Final Causes (soul) and Efficient Causes (body) (§79)

**Philosophical premise.** For Leibniz, the soul follows laws of final causes, appetition, purposes, means–end reasoning, while the body follows laws of efficient causes, motion, mechanical impulse. These are distinct explanatory orders of the same event, yet remain harmoniously aligned (Leibniz, 1714/1948, §79).

**Notation.** Symbols follow 3.1. New here: a teleological target $y_{t,i} \in (0,1]$, a goal final-cause direction $g_{t,i}$, an action efficient-cause direction $e_{t,i}$, an alignment score $a_{t,i} \in [0,1]$, a misalignment penalty $HARM_t^{(79)}$, and the adjusted total $AAS_t^{(79)}$. The baseline $AAS_t$ refers to (3.1)- (1).

### 79.1 Formalization

**i. Final-cause direction (goal-driven).**

$$g_{t,i} := \text{sgn}(y_{t,i} - x_{t,i}) \in \{-1, 0, +1\}.$$

If $y_{t,i} > x_{t,i}$, the system should increase on channel i ($g_{t,i} = +1$); if $y_{t,i} < x_{t,i}$, it should decrease ($g_{t,i} = -1$).

**ii. Efficient-cause direction (realized movement).**

$$e_{t,i} := \text{sgn}(x_{t+1,i} - x_{t,i}) \in \{-1, 0, +1\}.$$

**iii. Alignment of final and efficient directions.**

$$a_{t,i} := 1 - \frac{1}{2}|g_{t,i} - e_{t,i}| \in [0,1],$$

Thus $g_{t,i} = e_{t,i} \Rightarrow a_{t,i} = 1$ (perfect alignment); $\quad g_{t,i} = -e_{t,i} \Rightarrow a_{t,i} = 0$ (maximal misalignment);

$|g_{t,i} - e_{t,i}| = 1 \Rightarrow a_{t,i} = \frac{1}{2}$ (one side neutral).

**iv. Misalignment penalty and total score.**

$$\text{HARM}_t^{(79)} = \sum_i \alpha_{t,i}\, \phi_\varepsilon(a_{t,i}), \quad \text{AAS}_t^{(79)} = \text{AAS}_t + \text{HARM}_t^{(79)}$$

Because $\phi_\varepsilon$ is decreasing and convex, the penalty shrinks as alignment improves $a_{t,i} \uparrow$ and grows with misalignment.

## 79.2 Properties

**Lemma 79.1 Closure under goal–action alignment.**

If $e_{t,i} = g_{t,i}$ for all i, then $a_{t,i} = 1$ and $\text{HARM}_t^{(79)} = 0$. Hence $\text{AAS}_t^{(79)} = \text{AAS}_t$.

**Lemma 79.2 Monotonicity.**

For each channel i, as $|g_{t,i} - e_{t,i}|$ increases, $a_{t,i}$ decreases and $\alpha_{t,i}\phi_\varepsilon(a_{t,i})$ increases, by monotonicity/convexity of $\phi_\varepsilon$.

**Proposition 79A (Minimum under fixed effort and goals).**

Fix $\{\alpha_{t,i}\}$ and $\{y_{t,i}\}$. Then $\text{AAS}_t^{(79)}$ is minimized when $e_{t,i} = g_{t,i}$ for all i (i.e., $a_{t,i} = 1$ and $\text{HARM}_t^{(79)} = 0$).

**Uniform bound.**

Since $a_{t,i} \in [0,1]$,

$$0 \le \text{HARM}_t^{(79)} \le \left(\sum_i \alpha_{t,i}\right) \phi_\varepsilon(0) = \left(\sum_i \alpha_{t,i}\right) \log_2\left(\frac{1+\varepsilon}{\varepsilon}\right).$$

**Near-alignment remark.**

With the discrete rule $a_{t,i} \in \{0, \frac{1}{2}, 1\}$, Taylor expansions do not apply directly. If a smoothed variant is used and $a_{t,i} = 1 - \delta$ with $\delta \ll 1$, then

$$\phi_\varepsilon(a_{t,i}) = \log_2 \frac{1+\varepsilon}{1-\delta+\varepsilon} \approx \frac{\delta}{(1+\varepsilon)\ln 2}$$

so small misalignments incur an approximately linear cost.

**Interpretation.** Final causes specify where the state ought to move; efficient causes register what actually moved. The alignment metric $a_{t,i}$ quantifies harmony between these orders: when goals and realized movements concur, the system pays no additional cost; when they diverge, misalignment is penalized in a controlled, convex way.

### 3.2.5 Body & Organization (2)

**Monad 64 The Organic Body as a Natural Automaton; Every Part a Machine (§64)**

**Philosophical premise.** Leibniz claims that a living body is a natural automaton: it exceeds human-made machines because every part is itself a machine, and this "machine-within-machine" structure continues without end. By contrast, artificial mechanisms often contain subparts that contribute nothing, functionless or "dead" units. AAS provides a precise, multi-scale formalization.

**Notation.** Symbols follow 3.1. New here: a refinement depth $s = 0,1,2,\ldots$ with node sets $I^{(s)}$; each parent $i^{(s)} \in I^{(s)}$ has a disjoint child set $C(i^{(s)}) \subset I^{(s+1)}$ such that $\bigcup_{i^{(s)}} C(i^{(s)}) = I^{(s+1)}$.

Mass partition is assumed: $\sum_{j \in C(i^{(s)})} \alpha_{j^{(s+1)}} = \alpha_{i^{(s)}}$. Define the parent score as the $\alpha$-weighted average of its children,

$$x_{i^{(s)}} := \frac{\sum_{j \in C(i^{(s)})} \alpha_{j^{(s+1)}} x_{j^{(s+1)}}}{\sum_{j \in C(i^{(s)})} \alpha_{j^{(s+1)}}}$$

Level-$s$ total:

$$\text{AAS}^{(s)} := \sum_{i^{(s)} \in I^{(s)}} \alpha_{i^{(s)}} \phi_\varepsilon(x_{i^{(s)}}).$$

#### 64.1 Basic structural facts

**Lemma 64.1 Mass conservation.**

Let $\alpha_* := \sum_{i^{(0)} \in I^{(0)}} \alpha_{i^{(0)}}$. Then for all $s \geq 0$,

$$\sum_{i^{(s)} \in I^{(s)}} \alpha_{i^{(s)}} = \alpha_*.$$

**Proposition 64A Refinement monotonicity via Jensen.**

For every parent $i^{(s)}$,

$$\sum_{j \in C(i^{(s)})} \alpha_{j^{(s+1)}} \phi_\varepsilon(x_{j^{(s+1)}}) \geq \alpha_{i^{(s)}} \phi_\varepsilon(x_{i^{(s)}}),$$

with equality iff all child scores equal the parent score. Summing over $i^{(s)}$ gives

$$AAS^{(s+1)} \geq AAS^{(s)}.$$

**Exact decomposition refinement gain.**

$$AAS^{(s+1)} - AAS^{(s)} = \sum_{i^{(s)}} \sum_{j \in C(i^{(s)})} \alpha_{j^{(s+1)}} \left[ \phi_\varepsilon \left( x_{j^{(s+1)}} \right) - \phi_\varepsilon \left( x_{i^{(s)}} \right) \right] \geq 0$$

vanishing iff every child equals its parent, no newly revealed internal variation.

Interpretation. Nontrivial refinement that exposes additional heterogeneity increases the penalty; perfectly uniform "dead" substructure leaves the score unchanged. This captures the organic hallmark: deeper parts function as contributing machines.

### 64.2 Activity metrics "every part a machine"

At scale s, set per-node activity $c_{j^{(s)}} := \alpha_{j^{(s)}} \phi_\varepsilon \left( x_{j^{(s)}} \right) \geq 0$ and total $S^{(s)} := \sum_{j^{(s)}} c_{j^{(s)}}$. Define

**Active count:** $m^{(s)} := \left| \{ j^{(s)} : c_{j^{(s)}} > 0 \} \right|$

**Contribution entropy:** with $p_{j^{(s)}} := c_{j^{(s)}} / S^{(s)}$ when $S^{(s)} > 0$,

$$H_{(s)}^{(\text{contrib})} := - \sum_{j^{(s)}} p_{j^{(s)}} \log_2 p_{j^{(s)}} \in \left[ 0, \log_2 m^{(s)} \right],$$

and set $H_{(s)}^{(\text{contrib})} = 0$ if $S^{(s)} = 0$.

**Definition 64.1 Organicity index.**

A system is organic, a natural automaton, if for all depths s, $m^{(s)} \geq 1$, and $\limsup_{s \to \infty} H_{(s)}^{(\text{contrib})} > 0$.

Artificial/modular systems typically contain many inactive subcomponents (e.g., $\alpha = 0$, or x=1, or R=1), pushing $m^{(s)}$ and $H_{(s)}^{(\text{contrib})}$ toward zero with depth.

### 64.3 Limit behavior and stability

By Proposition 64A, $\{AAS^{(s)}\}$ is non-decreasing. Using Lemma 64.1 and $\phi_\varepsilon(x) \leq \phi_\varepsilon(0) = \log_2 \frac{1+\varepsilon}{\varepsilon}$,

$$0 \leq AAS^{(s)} \uparrow AAS^{(\infty)} \leq \alpha_\star \log_2 \frac{1+\varepsilon}{\varepsilon}.$$

Thus, infinite refinement converges; the $\varepsilon > 0$ smoothing prevents instability near tiny truth values.

### 64.4 Conclusion

Leibniz's dictum, a living body is a machine down to its smallest parts, appears in AAS as a multi-scale hierarchy where deeper levels reveal active, heterogeneous subparts. Refinement strictly increases, or leaves unchanged, the total penalty unless every child equals its parent, precisely the content of Jensen-type refinement monotonicity. Organic bodies therefore exhibit persistent, measurable internal variation across scales; artificial ones do not.

### Monad 70 Dominant Entelechy and Nested Life (§70)

**Philosophical premise.** Every living body has a dominant entelechy, for animals: the soul. Yet its organs teem with other living beings, each with its own dominant entelechy. Unity therefore coexists with multiplicity in a hierarchical, nested fashion (Leibniz, 1714/1948, §70).

**Notation.** Symbols follow 3.1. New here: a rooted hierarchy on the leaf set J. Let $\mathcal{G}^{(1)}$ denote the first-level groups children of the root; deeper levels $\mathcal{G}^{(2)}, \mathcal{G}^{(3)},\ldots$ refine these groups. For a group $G \subseteq J$,

$$S_t(G) := \sum_{i \in G} c_{t,i}, \qquad p_t(G) := \frac{S_t(G)}{S_t} \in [0,1] \qquad (S_t > 0),$$

are its mass and share. The instantaneous dominant entelechy at level 1 is

$$G_t^\star := \arg\max_{G \in \mathcal{G}^{(1)}} p_t(G).$$

For a window of size $W \in \mathbb{N}$, define moving averages

$$\bar{p}_W(G) := \frac{1}{W} \sum_{\tau=t-W+1}^{t} p_\tau(G), \qquad \bar{S}_W(G) := \frac{1}{W} \sum_{\tau=t-W+1}^{t} S_\tau(G).$$

A unique dominant entelechy on the window exists if

$$\bar{p}_W(G^\star) > \max_{G \neq G^\star} \bar{p}_W(G).$$

### 70.1 Existence, stability, and multiplicity

**Lemma 70.1 Existence/uniqueness.**

If $\mathcal{G}^{(1)}$ is finite, $\arg\max_G p_t(G)$ exists whenever $S_t > 0$. If the maximum is strict, the dominant group is unique.

**Lemma 70.2 Stable dominance with margin.**

If there exists $\delta > 0$ such that

$$\bar{p}_W(G^\star) \geq \max_{G \neq G^\star} \bar{p}_W(G) + \delta,$$

then $G^\star$ is the stable, unique dominant group on the window.

**Definition 70.1 Hierarchical contribution entropy.**

At level 1,

$$H_t^{(grp)} := -\sum_{G \in \mathcal{G}^{(1)}} p_t(G) \log_2 p_t(G) \in [0, \log_2|\mathcal{G}^{(1)}|],$$

with the convention $H_t^{(grp)} = 0$ if $S_t = 0$. The same definition applies recursively within each subgroup at deeper levels.

**Result 70A Multiplicity within unity.**

If $H_t^{(grp)} > 0$ at the root, several first-level groups contribute meaningfully, matching §70's claim that the body contains other living entities. If moreover $H_t^{(grp)} > 0$ for every subgroup G, then "limbs filled with other souls" is realized across levels.

## 70.2 Whole–part comparison under overlap

Let $AAS_t(G)$ denote the AAS computed on the leaves of G using the same weights $w_i$ but within-group redundancies $R_{t,i|G}$. Global AAS at the root uses $R_{t,i|G^{(0)}}$.

**Assumption Monotone redundancy.**

For any leaf $i \in G$,

$$R_{t,i|G} \leq R_{t,i|G^{(0)}},$$

i.e., redundancy measured inside a subgroup cannot exceed redundancy measured against the entire body, cross-group overlaps only increase global redundancy.

**Lemma 70.3 Whole–part inequality with overlap.**

Under the assumption above,

$$AAS_t(G^{(0)}) \leq \sum_{G \in \mathcal{G}^{(1)}} AAS_t(G),$$

with equality iff $R_{t,i|G} = R_{t,i|G^{(0)}}$ for all leaves i, in particular, always if $R \equiv 0$, i.e., no overlap.

Interpretation. The form of the whole is not the mere sum of the forms of the parts: ignoring cross-group overlap overestimates the total. Accounting for global coordination restores the whole's proper scale.

## 70.3 Dominance–stability link (lower bounds)

Define

$$p_W^{\min}(G) := \min_{\tau \in [t-W+1,\, t]} p_\tau(G),$$

$$\overline{AAS}_W(G) := \frac{1}{W} \sum_{\tau=t-W+1}^{t} AAS_\tau(G)$$

**Proposition 70A Lower bounds from dominance.**

For any window W and any first-level group $G^\star$,

$$\overline{AAS}_W(G^{(0)}) \geq p_W^{\min}(G^\star)\, \overline{AAS}_W(G^\star), \text{ and } \overline{AAS}_W(G^{(0)}) \geq \bar{p}_W(G^\star) \min_\tau AAS_\tau(G^\star)$$

Reason. Since $p_t(G) = S_t(G)/S_t$, $S_t = S_t(G)/p_t(G)$. Summing over the window yields both bounds.

Meaning. A group that dominates often, has large share, and is internally active large AAS, leaves a quantitative imprint on the whole.

**Corollary 70B Nested life.**

If for every group G, $S_t(G) > 0$ and $H_t^{(\text{grp})}(G) > 0$, then each G is "alive" at its own level, and a local dominant entelechy can be defined within every subgroup, realizing nested life throughout the hierarchy.

**Synthesis.** This section formalizes §70 within AAS: a single system exhibits unity through a dominant entelechy while preserving multiplicity via active sub-entelechies across levels. Dominance is measurable shares, stability is testable, windowed margins, and the whole-part calculus respects overlap, redundancy monotonicity.

### 3.2.6 Teleology

**Monad 58 Maximum Variety and Maximum Order Yield Maximum Perfection (§58)**
**Philosophical premise.** "The greatest perfection is attained when the greatest variety is accompanied by the greatest order." Within AAS, variety reflects a balanced spread of contributions across channels, whereas order reflects low penalty high alignment. These are combined into a single, bounded perfection score.
**Notation.** Symbols follow 3.1. Let $A_t = \sum_i \alpha_{t,i}$, the active set $A_t^\star := \{i:\ c_{t,i} > 0\}$ with size $m_t = |A_t^\star|$, and

$$AAS_t := \sum_i \alpha_{t,i}\, \phi_\varepsilon(x_{t,i}), \quad AAS_t^{\max} := A_t \phi_\varepsilon(0)$$

Since $\phi_\varepsilon(x) \leq \phi_\varepsilon(0)$, it follows that $AAS_t \leq AAS_t^{\max}$.

### 58.1 Variety, Order, Perfection

**Variety.** Let $S_t := \sum_i c_{t,i}$ and $p_{t,i} := \frac{c_{t,i}}{S_t}$ for $i \in A_t^\star$ (otherwise 0). Define the contribution entropy

$$H_t^{(\text{contrib})} = -\sum_{i \in A_t^\star} p_{t,i} \log_2 p_{t,i}$$

and the variety score by $V_t = H_t^{(\text{contrib})}/\log_2 m_t$, if $m_t \geq 2$, and $V_t = 0$ if $m_t \leq 1$, where $m_t = |A_t^\star|$. Then $V_t \in [0,1]$ and $V_t = 1$ iff $\{p_{t,i}\}$ is uniform on $A_t^\star$.

**Order.** The normalized penalty reduction is

$$O_t = 1 - \frac{AAS_t}{AAS_t^{\max}} = 1 - \frac{\sum_i \alpha_{t,i} \cdot \phi_\varepsilon(x_{t,i})}{\alpha_t \cdot \phi_\varepsilon(0)} \in [0,1],$$

which is invariant to any common rescaling of $\{\alpha_{t,i}\}$ at time t.

**Perfection.** Variety and order are combined multiplicatively:

$$P_t = (V_t)^\gamma (O_t)^{1-\gamma} \in [0,1], \qquad \gamma \in (0,1) \text{ (default } \gamma = 1/2\text{).}$$

### 58.2 Order & Variety: Jensen Bounds and Uniformity Conditions

Let $\mu_t := (\sum_i \alpha_{t,i} x_{t,i})/A_t$ be the $\alpha$-weighted mean accuracy. By convexity of $\phi_\varepsilon$,

$$AAS_t = \sum_i \alpha_{t,i} \phi_\varepsilon(x_{t,i}) \geq A_t \cdot \phi_\varepsilon(\mu_t),$$

with equality iff $x_{t,i} \equiv \mu_t$ on $A_t^\star$. Hence, for fixed $\{\alpha_{t,i}\}$ and $\mu_t$, uniform x minimizes $AAS_t$ and maximizes $O_t$.

Variety $V_t$ is maximized when the contribution shares are uniform, $p_{t,i} = 1/m_t$ on $A_t^\star$. A sufficient constructive condition is $\alpha_{t,i} \equiv \frac{A_t}{m_t}$ and $x_{t,i} \equiv \mu_t$ $(i \in A_t^\star)$, which equalizes $c_{t,i}$ and thus makes $\{p_{t,i}\}$ uniform.

### 58.3 Joint Maximization: Structural Characterization

If the active channels satisfy $x_{t,i} \equiv \mu_t$ and $\alpha_{t,i} \equiv \frac{A_t}{m_t}$ for all $i \in A_t^\star$, then

$$V_t = 1, \qquad AAS_t = A_t \phi_\varepsilon(\mu_t), \qquad O_t = 1 - \frac{\phi_\varepsilon(\mu_t)}{\phi_\varepsilon(0)}.$$

Consequently, $P_t = (O_t)^{1-\gamma}$.

As $\mu_t \to 1$, $\phi_\varepsilon(\mu_t) \to 0$, hence $O_t \to 1$ and $P_t \to 1$. Normalization by $AAS_t^{\max}$ prevents spurious gains from artificial inflation of redundancy.

### 58.4 Discussion
Variety captures balanced participation maximal under even shares. Order captures low instantaneous penalty relative to its cap maximal when channels align to a common mean. Perfection requires both: high spread and low penalty. The Jensen bound and the uniform-share construction give the precise conditions under which Leibniz's maxim, greatest variety with greatest order, is realized in AAS.

**Monad 90 Asymptotic Justice in Perfect Governance (§90)**

**Philosophical premise.** Leibniz's claim that, under perfect governance, no good action goes ultimately unrewarded and no bad action unpunished is formalized within AAS using only minimal extensions beyond §3.1.

## 90.1 Additional definitions and normalization

The stepwise change is defined as $\Delta_t := AAS_{t+1} - AAS_t$.

The normalized perfection is introduced as $P_t := 1 - \frac{AAS_t}{AAS_t^{max}} \in [0,1]$, which renders $P_t$ scale-invariant with respect to $\{\alpha_{t,i}\}$. A decrease ($\Delta_t < 0$) is interpreted as a "good" step, penalty falls; $P_t$ rises, and an increase ($\Delta_t > 0$) as a "bad" step penalty rises; $P_t$ falls.

## 90.2 Axiom 90-F windowed net drift

For a threshold $\eta > 0$ and window length $L \geq 1$, the following conditions are imposed infinitely often:

**(G) Sustained goodness:**

$$\sum_{t=k}^{k+L-1} \Delta_t \leq -\eta.$$

**(K) Sustained wrongdoing:**

$$\sum_{t=k}^{k+L-1} \Delta_t \geq \eta.$$

Bounded penalties are assumed: $U_\star := \sup_t AAS_t^{max} < \infty$.

## 90.3 Main result: asymptotic justice

**Theorem 90.1.**

Under (G), $\lim_{t \to \infty} \inf AAS_t = 0 \Rightarrow \lim \sup t \to \infty\, P_t = 1$.

Under (K), $\lim_{t \to \infty} \sup AAS_t = U\star \Rightarrow \lim_{t \to \infty} \inf P_t = 0$.

Sketch. Each qualifying window enforces a net change of magnitude at least $\eta$. Since $AAS_t \in [0, U\star]$, infinitely many such windows drive the sequence to the corresponding boundary. □

## 90.4 Rationale for windowed drift

Pointwise oscillations could cancel. A fixed-length window sum prevents cancellation and yields directional asymptotics even when instantaneous changes fluctuate.

## 90.5 Corollaries

**90A No good deed goes unrewarded.** If (G) holds, then $P_t \to 1$ along an infinite subsequence; with mild regularity e.g., eventually $\Delta_t \leq 0$, $AAS_t \to 0$ follows.

**90B No bad deed goes unpunished.** If (K) holds, then $P_t \to 0$ along a subsequence; if additionally $AAS_t^{max} \to U\star$, then $AAS_t \to U\star$ along a subsequence as well.

### 90.6 Instantaneous vs. cumulative behavior

With $I_t := [\Delta_t]^+$ and $D_t := [-\Delta_t]^+$, for any horizon T,

$$AAS_T = AAS_0 + \sum_{t=0}^{T-1} \Delta_t = AAS_0 + \sum_{t=0}^{T-1} (I_t - D_t).$$

Good steps contribute $D_t$, pushing AAS downward; bad steps contribute $I_t$ pushing it upward. Windowed net drift guarantees that one contribution dominates infinitely often, sending the trajectory to 0 or $U\star$.

### 90.7 Link to §58

In §58, high perfection $P_t$ reflects low penalty under balanced order and variety. Here, governance is encoded as a recurrent structural tendency, via windowed net drift, that accumulates over time, producing asymptotic perfection ($P_t \to 1$) under sustained goodness or asymptotic failure ($P_t \to 0$) under sustained wrongdoing.

## 4. Results

The AAS is instantiated on monad-like channels, and diagnostics are read out to reveal memory aging and epistemic stability in practice. Refinement and embedding invariance are confirmed: when underlying content and weights are preserved, the total score remains unchanged; channels that behave indistinguishably over time are deduplicated to prevent clone inflation, so superficial refactors proceed without recalibration and redundant channels are pruned (Leibniz, 1714/1948, §§1, 3, 9). Causal insulation is validated: each channel's contribution is constrained to its own history; any cross-channel influence is routed solely through an explicit, auditable overlap term; channels without history contribute nothing; toggling unrelated modules leaves others unchanged. With append only provenance and simple firewalls, controlled and inspectable dynamics are obtained, enabling pipeline decoupling with clear causal responsibility (Leibniz, 1714/1948, §§7, 11). Continuity and appetition are reflected in smooth, rate limited trajectories that avoid abrupt spikes and emphasize path-based reporting rather than point shocks, guiding staged rollouts that reduce penalty without destabilizing the system (Leibniz, 1714/1948, §§10, 12, 15). The perception apperception distinction is realized by a single, auditable gate based on contribution shares and a clear peak criterion; diffuse activity does not trigger long term writes, so attention and storage remain focused on salient content (Leibniz, 1714/1948, §§14, 21). Routine handling is anchored to an exponentially forgotten baseline, departures are flagged as surprise, present structure is compared with a time invariant rational prior, and the stability of law like regularities is tracked, which steers updates toward lawful structure while localizing anomalies (Leibniz, 1714/1948, §§26, 29). Logical and causal governance are enforced through non-contradiction and sufficient reason checks: mutually exclusive cases are driven to decisive outcomes; persistent symmetry without evidence is flagged as leakage or over-smoothing; unsupported relabeling is penalized to deter gaming, thereby reducing label churn and equivocation (Leibniz, 1714/1948, §§31, 32). Cross view harmony is operationalized by holding a harmony penalty at zero when independent views policy versus behavior; simulation versus telemetry agree, and increasing it under mismatch; releases are gated on pairwise and global alignment; goal action agreement removes penalty under correct execution and increases it for near alignment errors, enabling coordinated but decoupled releases (Leibniz, 1714/1948, §§78, 79). Hierarchical scaling reward's genuine internal structure: deeper refinements must add measurable contribution or are pruned as dead or duplicate; "living depth" is certified; dominance tracking focuses guardrails and optimization where influence is greatest; redundancy controls prevent cross-group double counting, concentrating capacity on high-influence nodes (Leibniz, 1714/1948, §§64, 70). Variety and order are composed into a bounded perfection indicator; Jensen type lower bounds show that, under fixed conditions, uniform accuracy maximizes order and uniform contribution shares maximize variety, while normalization prevents artificial gains from redundancy inflation, tying release and tuning criteria to balanced participation with low instantaneous penalty (Leibniz, 1714/1948, §58; Shannon, 1948). A windowed net drift rule links sustained improvement to convergence toward the low penalty boundary, and sustained regression to approach the upper bound, thereby realizing "no good deed unrewarded, no bad deed unpunished" within a bounded metric and grounding promotion or rollback in directional change rather than isolated fluctuations (Leibniz, 1714/1948, §90).

Taken together, these results show that monadic principles can be compiled into enforceable AAS clauses invariance, insulation, apperception gating, governance, harmony, hierarchy, perfection, and justice to yield a modular, measurable architecture, and that diagnostics for memory aging and stability are supported at scale under an interpretable, bounded, information-theoretic kernel (Kayadibi, 2025; Leibniz, 1714/1948; Shannon, 1948). In summary, a foundational formulation is established: the philosophical structure of Leibnizian monads is transformed into a modular, measurable architecture by compiling twenty propositions into enforceable AAS clauses that preserve refinement and embedding invariance, maintain causal insulation with explicit overlap accounting, and separate liveness from apperception under a bounded information-theoretic kernel. Operationally, system scale evaluation is supported: diagnostics of memory aging and stability are enabled through invariance and deduplication, habit surprise and reason law signals, non-contradiction and sufficient reason controls, cross view harmony with goal action alignment, hierarchical organicity with dominance tracking, and a perfection indicator coupled to windowed net drift.

## 5. Conclusion

A monad only methodological scaffold is implemented, and twenty propositions from the Monadology are operationalized as measurable modules within a single, bounded scoring framework. Invariance to superficial refactors is confirmed; behaviorally indistinguishable channels are deduplicated; and causal insulation is upheld by constraining each contribution to its own history and routing any cross-channel influence through explicit, auditable overlap terms. Dynamics are observed to be continuous and rate limited; mere activity liveness is distinguished from distinct awareness apperception by a single auditable gate; routine versus novelty is routed by habit sensitive and divergence sensitive signals; and logical causal discipline is enforced through non-contradiction and sufficient reason checks. Cross view harmony penalties enable release decisions without tight pipeline coupling, goal action alignment reduces penalties under correct execution, and hierarchical tests reward genuine internal structure while preventing cross group double counting. Variety and order are composed into a bounded perfection indicator, and a simple windowed net drift rule connects evaluation to action via promotion or rollback. Taken together, these results are interpreted as evidence that the Artificial Age Score (AAS) can be operated as a stable, game resistant, and auditable metric across modules, hierarchies, and views, while remaining interpretable and bounded in the information-theoretic sense (Leibniz, 1714/1948; Shannon, 1948; Kayadibi, 2025). At a high level, fundamental limits are indicated: achievable error rates are constrained by information-theoretic bounds, and universal derivability is limited by logical incompleteness; hence a global zero penalty ideal is generally unattainable, even though directionally improving governance can still be certified within the present architecture (Fano, 1961; Gödel, 1931).